\documentclass[11pt]{article}
\usepackage{coling2016}
\usepackage{color}
\usepackage{times}
\usepackage{latexsym}
\usepackage{amsmath,amssymb,amsthm}
\usepackage{graphicx}
\usepackage{tikz}
\usetikzlibrary{arrows,shapes,positioning,backgrounds}

\title{Language classification from bilingual word embedding graphs}

\author{Steffen Eger \\
  UKP Lab \\
	    TU Darmstadt\\
	    64289 Darmstadt\\
	  \And
	Armin Hoenen\\
        Text Technology Lab \\
  	Goethe University \\ 
  	60325 Frankfurt am Main \\
  \And Alexander Mehler\\
  Text Technology Lab \\
  	Goethe University \\ 
  	60325 Frankfurt am Main \\
}

\date{}

\begin{document}

\maketitle

\begin{abstract}
We study the role of the second language in bilingual word embeddings
in 
{monolingual semantic evaluation tasks}.
We find strongly and weakly positive correlations between down-stream
task performance and second language similarity to the target
language. 
Additionally, 
we show how bilingual word embeddings can be employed for the task of
semantic {language classification} and that 
joint semantic spaces 
vary in meaningful ways across second languages. Our results support 
the 
hypothesis that 
semantic language similarity is influenced by both
structural similarity as well as
geography/contact.  
\end{abstract}

\section{Introduction}\label{sec:introduction}
Word embeddings derived from context-predicting neural
network architectures have become the state-of-the-art in
distributional semantics modeling \cite{Baroni:2014}. Given the success
of these models and the ensuing hype, several extensions over the
standard paradigm 
\cite{Bengio:2003,Collobert:2008,Mikolov:2013,Pennington:2014} have
been suggested, such as 
retro-fitting 
word vectors to semantic knowledge-bases \cite{Faruqui:2015},
multi-sense \cite{Huang:2012,Neelakantan:2014}, 
and multi-lingual word vectors
\cite{Klementiev:2012,Faruqui:2014,Hermann:2014,Chandar:2014,Lu:2015,Gouws:2015,Gouws:2015:NAACL-HLT,Huang:2015,Suster:2016}. 

\begin{figure}[!htb]
\begin{center}
\input{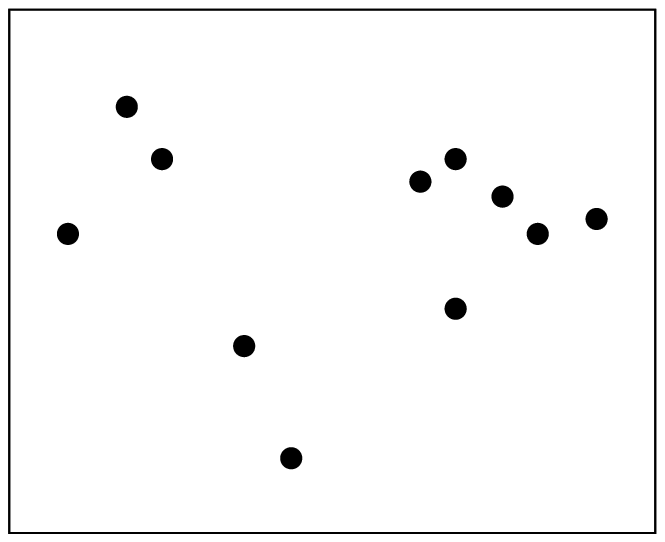}
\input{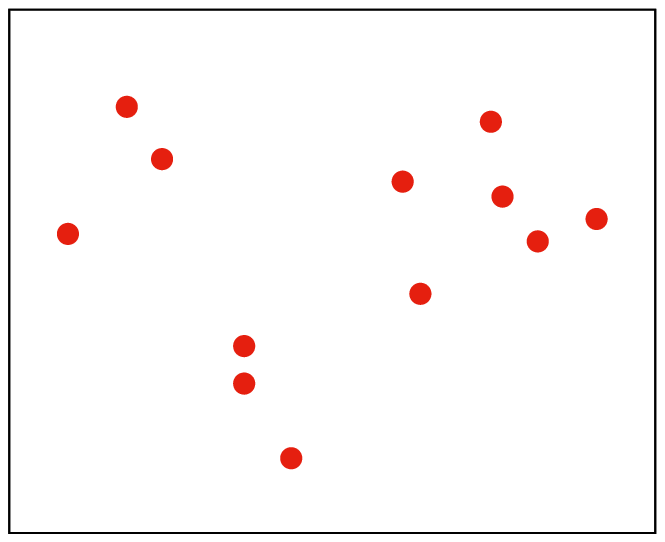}
\input{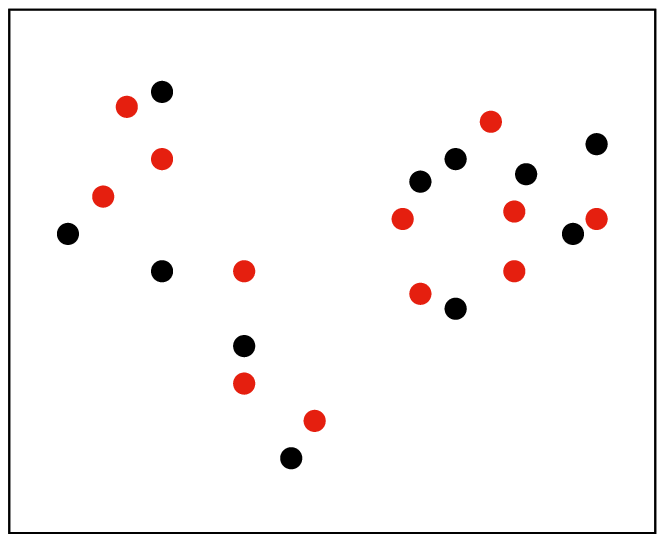}
\input{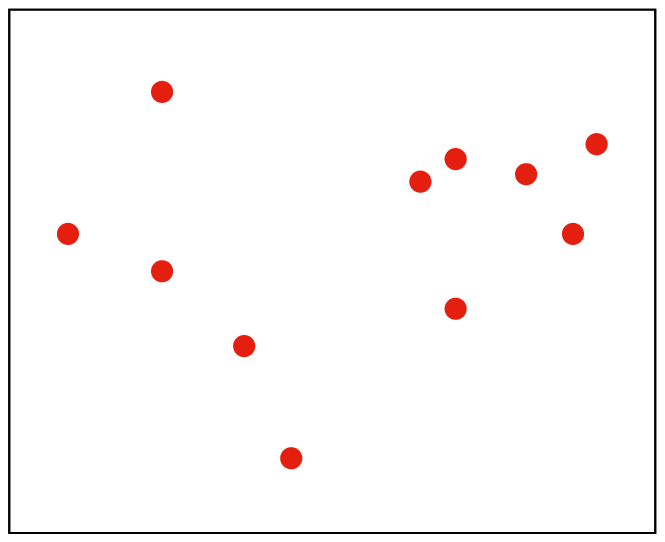}
\end{center}
\caption{Monolingual embeddings (top left and top right) have been
shown to capture semantic (as well as syntactic) properties of
languages, here exemplarily: $p=$ English and $\ell=$ Latin. Bottom
left: The (idealized) goal of crosslingual embeddings is to capture
these relationships across two or more languages. Bottom right: After
projection in a joint semantic space, semantic (as well as syntactic)
properties of words in language $p$ have adapted to those of language
$\ell$. Note, in particular, the movement of \emph{man} in these
idealized plots, i.e., the different positions of \emph{man} in top
left vs.\ bottom right.  
}
\label{fig:gouws}
\end{figure}

The 
models 
underlying the 
latter paradigm, which we focus on in the current work, project word
vectors of two (or multiple) languages into  
a joint semantic space, thereby allowing to evaluate semantic
similarity of words from different languages; see
Figure \ref{fig:gouws} for an illustration. Moreover, the resulting
word vectors have been shown to produce 
on-par or better performance even in a
\emph{monolingual} setting, e.g., when using them for measuring
semantic similarity in one of the two languages involved \cite{Faruqui:2014}. 

While
multilingual word vectors have been evaluated with respect to
intrinsic parameters such as embedding dimensionality, empirical work
on another aspect appears to be lacking: the second language
involved. For example, it might be the case that projecting two
languages with very different lexical semantic associations in a joint
embedding space inherently deteriorates \emph{monolingual} embeddings
as measured by performance on an intrinsic {monolingual}
semantic evaluation task, 
relative to a setting in which the two
languages have very similar lexical semantic
associations. 
To illustrate, the classical Latin word \textcolor{black}{\emph{vir}} is sometimes translated in
English as both \textcolor{black}{`man' and 
`warrior',} suggesting a semantic connotation, in Latin, that is
putatively lacking 
in English. Hence, projecting English and Latin in a joint semantic
space may invoke semantic relations that are misleading for an
English evaluation task. 
Alternatively, it may be argued that heterogeneity in semantics
between the two languages involved is
beneficial for monolingual evaluation tasks in the same way
that uncorrelatedness in classifiers helps in combining them.

Here, we study two questions (\textbf{main contributions}). On the one
hand, we are interested in the 
effect of language similarity on bilingual word embeddings in a
(Q1) \textbf{monolingual (intrinsic) semantic evaluation task}. Thus,
our first 
question is: \textcolor{black}{how does the performance of bilingual word embeddings in
monolingual semantic evaluation tasks depend on the second language
involved?}\footnote{Our initial expectation was
that bilingual word embeddings lead to better results in
monolingual settings, at least for some second languages. However,
this was not confirmed in any of our experiments. 
This may be related to our
(small) data set sizes (see Section \ref{sec:data}) or to other
factors, but has no concern for the question (Q1) we are
investigating.} 
Secondly, we ask how
bilingual word embeddings can be employed for the task of
semantic (Q2) \textbf{language classification}.   
Our approach here is simple: we project languages onto a common pivot
language $p$ so as to make them comparable. We directly use bilingual word
embeddings for this. More precisely, we first project languages 
$\ell$ in a common semantic space with the pivot $p$ by means of
bilingual word embedding methods. Subsequently, we ignore language $\ell$ words
in the joint space. 
Semantic distance measurement between
languages then amounts to comparison of graphs that have the same
nodes --- pivot language words --- and different edge weights --- semantic
similarity scores between pivot language words based on 
bilingual embeddings that vary as a function of the second language
$\ell$ involved. This core idea is illustrated in
Figures \ref{fig:gouws} and \ref{fig:ill}.   

We show that 
joint semantic spaces induced by bilingual word embeddings
vary in meaningful ways across second languages. 
Moreover, our results support the hypothesis that 
semantic language similarity is 
influenced by both genealogical language similarity and by
aspects of language contact.  

This work is structured as follows. Section \ref{sec:model} introduces
our approach of constructing graphs from bilingual word embeddings and
its relation to the two questions outlined. Section \ref{sec:data}
describes our data, which is based on the Europarl
corpus \cite{Koehn:2005}. Section \ref{sec:experiments} details our
experiments, which we discuss in Section \ref{sec:discussion}. We
relate to previous work in Section \ref{sec:related} and conclude in
Section \ref{sec:conclusion}.

\section{Model}\label{sec:model}
In this section, we formally outline our approach. 

Given $N+1$ languages, choose one of them, $p$, as \emph{pivot
language}. Construct $N$ weighted networks
$G_{\ell}^{(p)}=(V^{(p)},E^{(p)},\mathsf{w}_{\ell}^{(p)})$ as follows: nodes $V^{(p)}$
are the words of language $p$, graphs are fully connected, i.e.,
$E^{(p)}=V^{(p)}\times V^{(p)}$, 
and edge weights are 
$\mathsf{w}_{\ell}^{(p)}(u,v)=\mathsf{sim}(\mathbf{u}_{p,\ell},\mathbf{v}_{p,\ell})$. 
The similarity function 
$\mathsf{sim}$ is, e.g., cosine similarity, and
$\mathbf{u}_{p,\ell},\mathbf{v}_{p,\ell}\in\mathbb{R}^d$ are
\emph{bilingual} word embeddings of words $u$ and $v$, respectively,
derived from any suitable 
method (see below). 
Here, $\ell$ ranges over the $N$ \emph{second} 
\emph{languages}. 

For (Q1) \textbf{monolingual semantic evaluation} in language 
$p$, 
choose $p$ as pivot and consider $G_{\ell}^{(p)}$ for varying
second languages $\ell$. 
 We can then evaluate semantic similarity between
two language $p$ words $u$ and $v$ by querying the edge weight
$\mathsf{w}_{\ell}^{(p)}(u,v)$. This is the classical situation of (intrinsic) monolingual
evaluation of bilingual word embeddings. 

For (Q2) \textbf{language classification}, we compare the graphs $G_{\ell}^{(p)}$
across all second languages $\ell$, and a fixed pivot $p$. 
Here, we have many choices how to realize distance measures between
graphs, such as which metric we use \textcolor{black}{and} at which
level we compare  
graphs \cite{Bunke:1998,Rothe:2014}.  
We choose the following: 
we first represent 
each node (pivot language word) in a graph as the vector of distances
to all other nodes. 
That is, 
to $u\in V[G_{\ell}^{(p)}]=V^{(p)}$ we assign the
vector 
$\mathbf{r}_u^{(\ell)}=\bigl(\mathsf{w}_\ell^{(p)}(u,v)\bigr)_{v\in
V^{(p)}}$. The distance 
$d(G_\ell^{(p)},G_{\ell'}^{(p)})$ 
between two graphs 
$G_\ell^{(p)}$ and $G_{\ell'}^{(p)}$  is then defined as the average
distance (Euclidean norm) of the so represented nodes in the
graphs: $d(G_\ell^{(p)},G_{\ell'}^{(p)})=\frac{1}{|V^{(p)}|}\sum_{u\in
V^{(p)}}||\mathbf{r}_u^{(\ell)}-\mathbf{r}_u^{(\ell')}||$. Finally, we define
the \emph{(syntacto-)semantic distance} 
$D(\ell,\ell')$ of two languages 
$\ell$ and $\ell'$ as the average graph distance over all $N-1$ pivots
($l$ and $l'$ excluded): 
\begin{equation}
\begin{aligned}
  D(\ell,\ell') = \frac{1}{N-1}\sum_{\tilde{p}}
  d(G_{\ell}^{(\tilde{p})},G_{\ell'}^{(\tilde{p})}). 
\end{aligned}
\label{eq:one}
\end{equation}
By summing over pivots, we effectively `integrate out' the
influence of the pivot language, leading to a `pivot independent'
language distance calculation. In addition, this ensures that the
distance matrix $D$ encompasses all languages, including all possible
pivots.   

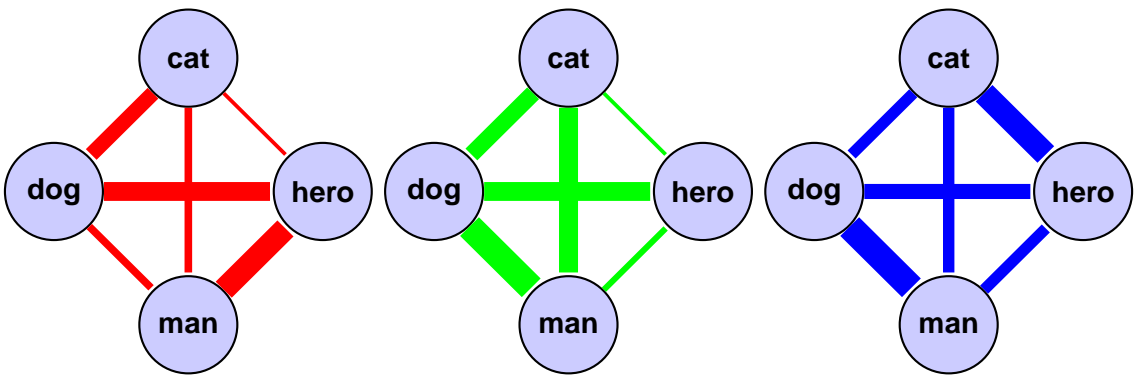
\begin{figure}[!htb]
        \begin{center}
        \begin{tikzpicture}[-,>=stealth',shorten >=1pt,auto,node distance=2.5cm,
  thick,main node/.style={circle,fill=blue!20,draw,minimum width={width("hamster")+2pt},font=\sffamily\bfseries}]

  \begin{scope}
  \node[main node] (1) {cat};
  \node[main node] (2) [below left of=1] {dog};
  \node[main node] (3) [below right of=2] {man};
  \node[main node] (4) [below right of=1] {hero};


  \tikzset{Friend/.style   = {
                                 double          = red,
                                 double distance = 1pt}}
  \tikzset{Enemy/.style   = {
                                 double          = red,
                                 double distance = 1pt}}

  \draw[line width=2mm, draw=red](1) to (2); 
  \draw[line width=1mm, draw=red](1) to (3);
  \draw[line width=0.5mm, draw=red](1) to (4);
  \draw[line width=1mm, draw=red](2) to (3);
  \draw[line width=2.5mm, draw=red](2) to (4);
  \draw[line width=3mm, draw=red](3) to (4);


\end{scope}
  \begin{scope}[xshift=5cm]
  \node[main node] (1) {cat};
  \node[main node] (2) [below left of=1] {dog};
  \node[main node] (3) [below right of=2] {man};
  \node[main node] (4) [below right of=1] {hero};


  \tikzset{Friend/.style   = {
                                 double          = green,
                                 double distance = 1pt}}
  \tikzset{Enemy/.style   = {
                                 double          = green,
                                 double distance = 1pt}}

  \draw[line width=2.2mm, draw=green](1) to (2); 
  \draw[line width=2.5mm, draw=green](1) to (3);
  \draw[line width=0.5mm, draw=green](1) to (4);
  \draw[line width=3.5mm, draw=green](2) to (3);
  \draw[line width=2.5mm, draw=green](2) to (4);
  \draw[line width=0.8mm, draw=green](3) to (4);


\end{scope}
  \begin{scope}[xshift=10cm]
  \node[main node] (1) {cat};
  \node[main node] (2) [below left of=1] {dog};
  \node[main node] (3) [below right of=2] {man};
  \node[main node] (4) [below right of=1] {hero};


  \tikzset{Friend/.style   = {
                                 double          = blue,
                                 double distance = 1pt}}
  \tikzset{Enemy/.style   = {
                                 double          = green,
                                 double distance = 1pt}}

  \draw[line width=1.5mm, draw=blue](1) to (2); 
  \draw[line width=1.5mm, draw=blue](1) to (3);
  \draw[line width=3.1mm, draw=blue](1) to (4);
  \draw[line width=3.5mm, draw=blue](2) to (3);
  \draw[line width=2mm, draw=blue](2) to (4);
  \draw[line width=1.5mm, draw=blue](3) to (4);


\end{scope}
\end{tikzpicture}
        \end{center}
        \caption{Schematic illustration of our approach. Repeating the ``four-stage'' process illustrated in Figure \ref{fig:gouws} for three different languages $\ell$ (marked by different colors) and the same pivot $p$. Edge strengths between pivot language words indicate their semantic similarity as measured by cosine distances in semantic spaces as in Figure \ref{fig:gouws} bottom right.  
        }
        \label{fig:ill}
\end{figure}

Figure \ref{fig:ill} illustrates our idea of projecting semantic
spaces of different languages onto a common pivot.  

\textbf{Bilingual embedding models}:
We consider two approaches to constructing bilingual word embeddings. 
The first is the canonical
correlation analysis (\textbf{CCA}) approach suggested in
\newcite{Faruqui:2014}. This 
takes \emph{independently} constructed word vectors from two different
languages and projects them onto a common vector space such
that (one-best) translation pairs, as determined by automatic word
alignments, 
are maximally linearly correlated. CCA relies on word level alignments
and we use cdec for this \cite{Dyer:2010}. 

The second approach we
employ is called BilBOWA (\textbf{BBA}) \cite{Gouws:2015}. Rather than
separately training word vectors for two languages and subsequently
enforcing cross-lingual constraints, this model \emph{jointly} optimizes
monolingual and cross-lingual objectives similarly as in
\newcite{Klementiev:2012}: 
\begin{align*}
  \mathcal{L} = \sum_{\ell \in \{e,f\}}\sum_{w,h\in
    \mathfrak{D}^\ell} \mathcal{L}^\ell(w,h;\theta^\ell)+\lambda\Omega(\theta^e,\theta^f)
\end{align*}
is minimized, where $w$ and $h$ are target words and their contexts,
respectively, and $\theta^e,\theta^f$ are embedding parameters for two
languages. The terms $\mathcal{L}^\ell$ encode the monolingual
constraints and the term $\Omega(\theta^e,\theta^f)$ encodes the 
cross-lingual constraints, enforcing similar words across languages
(obtained from \emph{sentence aligned} data) to have similar
embeddings. 

\section{Data}\label{sec:data}
For our experiments, we use the Wikipedia extracts available from
\newcite{Al-Rfou:2013}\footnote{{https://sites.google.com/site/rmyeid/projects/polyglot}.}
as monolingual data and Europarl \cite{Koehn:2005} as bilingual 
database. We consider two settings, one in which we take all 21 (\texttt{All21})
languages available in Europarl and one in which we focus on the 10
(\texttt{Big10}) 
largest languages. 
These languages are bg, cs, \textbf{da}, \textbf{de} , el,
\textbf{en}, \textbf{es}, et, \textbf{fi}, \textbf{fr}, hu,
\textbf{it},  
lt, lv, \textbf{nl}, pl, \textbf{pt}, ro, sk, sl, \textbf{sv}
(\texttt{Big10} languages 
highlighted). 
To induce a comparable setting, we extract
in the \texttt{All21} setup: 
195,842
parallel sentences from Europarl 
and roughly 835K (randomly extracted) sentences from
Wikipedia for each of the 21 languages. In the \texttt{Big10} setup,
we extract 1,098,897 parallel sentences from Europarl and 2,540K
sentences from 
Wikipedia for each of the 10 languages involved. 
We note that the above numbers are determined by the minimum available
for the respective two sets of languages in the Europarl and Wikipedia
data, respectively. 
As preprocessing, we
tokenize all sentences in all datasets and we lower-case all words.

\section{Experiments}\label{sec:experiments}
We first train $d=$ 200 dimensional skip-gram word2vec
vectors \cite{Mikolov:2013}\footnote{All other  
parameters set to default values.} on the union of the
Europarl and Wikipedia data for each language in the respective
\texttt{All21} and \texttt{Big10} setting. For CCA, we then obtain
bilingual embeddings for each possible combination $(\ell,\ell')$ of
languages in 
each of the two setups, by projecting these vectors in a joint space
via word alignments obtained on the respective Europarl data
pair.
For BBA, we use the monolingual Wikipedias of $\ell$ and
$\ell'$ for the 
monolinugal constraints, and the Europarl sentence alignments of
$\ell$ and $\ell'$ for the bilingual constraints. 
We only
consider words that occur at least 100 times in the respective data
sets. 

\subsection{Monolingual semantic task (Q1)}
We first evaluate the obtained BBA and CCA embedding vectors 
on monolingual $p=$ English evaluation tasks, for varying second
language $\ell$. 
The tasks we consider are WS353 \cite{Finkelstein:2002}, MTurk287
\cite{Radinsky:2011}, MTurk771,\footnote{http://www2.mta.ac.il/~gideon/mturk771.html} SimLex999 \cite{Hill:2015}, and MEN 
\cite{Bruni:2014}, which are standard semantic
similarity datasets for English, documented in
an array of previous research. 
In addition, we include the SimLex999-de and SimLex999-it
\cite{Leviant:2015} for $p=$ German and $p=$ Italian, respectively.
In each task, the goal is to determine the semantic similarity between
two language $p$ words, such as \emph{dog} and \emph{cat} (when $p= $
English). 
For the tasks, we indicate average Spearman
correlation coefficients $\delta=\delta_{p,\ell}$ between 
            \begin{itemize}
            \item the predicted semantic
similarity --- measured in cosine similarity --- 
between respective language $p$ word pair vectors obtained when projecting $p$ and $\ell$ in
a joint embedding space, 
and 
    \item the
human gold standard (i.e., human judges have assigned semantic
similarity scores to word pairs such as \emph{dog,cat}).
\end{itemize}

Table \ref{table:ws353} below 
exemplarily lists results for MTurk-287, for which $p=$ English. We
notice two trends. First, for BBA, results can roughly be partitioned
into three 
classes. The languages pt, es, fr, it have best performances as second
languages with $\delta$ values between 54\% and close to 60\%; the
next group consists of da, nl, ro, de, el, bg, sv, sl,
cs with values of around 50\%; finally, fi, pl, hu, lv, lt, sk, et
perform worst as second languages with $\delta$ values of around
47\%. So, for BBA, the choice of second language has evidently a
considerable 
effect in that there is {\raise.17ex\hbox{$\scriptstyle\sim$}}26\%
difference in performance between 
best second language,  
$\ell=$ it, and worst second languages, $\ell=$ pl/sk/et. Moreover, it
is apparently 
better to choose (semantically) similar languages --- with reference
to the target language $p=$ English --- as second language
in this case. Secondly, for CCA, variation in results
is much less pronounced. For example, the best second languages,
et/lv, are 
just roughly 5.5\% better than the worst second language,
lt. Moreover, it is not evident, on first view, that performance
results depend on language similarity in this case.\footnote{As
  further results, we note \emph{en passant}: CCA performed typically
  better than BBA, particularly in three --- MEN, WS353, SimLex999 --- out of
  our 
  five English datasets as well as the non-English datasets. This
  could be due to the fact that we trained the 
  vectors for the skip-gram model --- the monolingual vectors that
  form the basis for CCA --- on the union of Europarl and Wikipedia,
  while BBA used only Wikipedia as a monolingual basis. 
  Other explanations could be the particular default hyperparameters
  chosen, which may have coincidentally favored CCA, or the fact that
  CCA uses only 1-best word alignments for projection; see
  Section \ref{sec:discussion} for further discussions. 
  Moreover, in no
  case did we find that either BBA or CCA outperformed the purely
  monolingually constructed skip-gram vectors on the English
  evaluation task. On the one hand, this may be due to our rather
  small bilingual 
databases --- containing just roughly 200K and 1,000K parallel
sentences. On the other hand, while this finding is partly at odds with
\newcite{Faruqui:2014}, who report large improvements for bilingual
word vectors over monolingual ones in some settings, it is (more) in
congruence with  
\newcite{Lu:2015} and \newcite{Huang:2015}.}
\begin{table}[!htb]
\begin{center}
\begin{tabular}{|ccc|} \hline
   & BBA & CCA \\ \hline
 pt & 56.54 & 57.48\\
 es & 54.87 & 56.76\\
 fr & 54.48 & 56.76\\
 it & 59.70 & 57.12\\
 da & 50.49 & 56.49\\
 nl & 49.94 & 57.49\\
 ro & 51.44 & 58.10\\
 de & 50.08 & 58.24\\
 el & 51.23 & 56.66\\
 bg & 49.90 & 57.04\\
 \hline
\end{tabular}
\begin{tabular}{|ccc|} \hline
   & BBA & CCA \\ \hline
 sv & 50.02 & 56.06\\
 fi & 47.41 & 56.76\\
 pl & 47.12 & 56.47\\
 cs & 49.94 & 56.74\\
 sl & 52.96 & 57.05\\
 hu & 48.84 & 56.46\\
 lv & 47.55 & 58.81\\
 lt & 47.49 & 55.66\\
 sk & 47.22 & 57.17\\
 et & 47.26 & 58.75\\
 \hline
\end{tabular}
\end{center}
\caption{Correlation coefficients $\delta=\delta_{p,\ell}$ in \% on
 MTurk-287 for BBA 
 and CCA methods, respectively, for various second languages $\ell$. Second
 languages ordered by semantic similarity to $p=$ English, as determined by
 Eq.~\eqref{eq:one}; see \S\ref{sec:lang_sim} for specifics.}
\label{table:ws353}
\end{table}

To quantify this, we systematically compute correlation coefficients
$\tau$ between the correlation coefficients $\delta=\delta_{p,\ell}$
and the language distance values $D(p,\ell)$ from Eq.~\eqref{eq:one}
(see \S\ref{sec:lang_sim} for specifics on $D(p,\ell)$). 
Table \ref{table:corr} shows that, indeed, 
monolingual semantic evaluation performance is consistently positively 
correlated 
with (semantic) language similarity for BBA. 
In contrast, for CCA, correlation is
positive in eight cases and negative in six cases; moreover, coefficients
are significant in only two cases. Overall, there is a strongly
positive average correlation for BBA (75.75\%) and a (very) weakly
positive one for CCA (10.04\%). 
\begin{table}[!htb]
\begin{center}
\begin{tabular}{|l|lll|} \hline
 $p=$&   & BBA & CCA \\ \hline\hline
 en &WS353-\texttt{All21} & 63.75** & -6.16\\
 &WS353-\texttt{Big10} & 93.33** & 58.33$\dagger$\\
 &MTurk287-\texttt{All21} & 80.75***& 5.11\\
 &MTurk287-\texttt{Big10} & 88.33** & -21.66\\
 &MTurk771-\texttt{All21} & 74.28*** & -19.24 \\
 &MTurk771-\texttt{Big10} & 93.33*** & 11.66\\
 &SimLex999-\texttt{All21} & 83.60*** & 11.57 \\
 &SimLex999-\texttt{Big10} & 73.33* & -20.00 \\
 &MEN-\texttt{All21} & 70.82*** & -11.27\\
 &MEN-\texttt{Big10} & 94.99*** & 41.66 \\ \hline
 de &SimLex999-de-\texttt{All21} & 60.45** & 10.07 \\ 
 &SimLex999-de-\texttt{Big10} & 73.33* & -31.66 \\ \hline
 it &SimLex999-it-\texttt{All21} & 48.57* & 73.83*** \\ 
 &SimLex999-it-\texttt{Big10} &  61.66$\dagger$ & 38.33 \\ \hline\hline
 &Avg. & 75.75 & 10.04 \\ \hline
\end{tabular}
\caption{Correlation $\tau$, in \%, between language similarity and
 monolingual semantic
 evaluation performance. For example, on WS353 in the \texttt{Big10}
 setup, the more a language, say $\ell=$ French, is (semantically) similar to
 $p=$ English, the more is it likely that correlations $\delta_{p,\ell}$ 
 are large, when word pair similarity of
 $p=$ English words is measured from 
 embedding vectors
 that have been projected in a joint French-English semantic embedding
 space. More precisely, the exact correlation values are 93.33\%
 and 58.33\%, 
 respectively, depending on whether vectors have been projected via
 BBA or CCA. 
{`***' means significant at the 0.1\% level; `**' at the 1\%
  level, `*' at the 5\% level, '$\dagger$' at the 10\% level.}}
\label{table:corr}
\end{center}
\end{table}

\subsection{Language classification (Q2)}\label{sec:lang_sim}
Finally, we perform language classification on the graphs $G_\ell^{(p)}$
as indicated in \S\ref{sec:model}. Since we use two different methods
for inducing bilingual word embeddings, we obtain two distance
matrices.\footnote{For \texttt{All21}, these two distance matrices
  have a correlation of close to 70\% (Mantel test), and of 73\% for
  \texttt{Big10}. Hence, overall, semantic language classification
  results 
  produced by either of the two methods alone --- BBA or CCA --- are
  expected to be very similar. 
  } 
Figure \ref{fig:all21} below  
shows a two-dimensional representation of all 21 languages 
obtained from \emph{averaging} the BBA and CCA distance matrices 
in the \texttt{All21} setup, 
together with a $k$-means cluster assignment for $k=6$. We note a grouping
together of es, pt, fr, en, it; \textcolor{blue}{nl, da, de, sv}; \textcolor{cyan}{fi, et}; \textcolor{green}{ro, bg, el};
\textcolor{red}{hu, pl, cs, sk, sl}; and \textcolor{magenta}{lt, lv}. In particular, $\{ \text{es, pt, fr,
it, en}\}$ appear to form a homogeneous group with, consequently, 
similar semantic 
associations, as captured by word embeddings. 
\begin{figure}[!htb]
\begin{center}
  \includegraphics[scale=0.60]{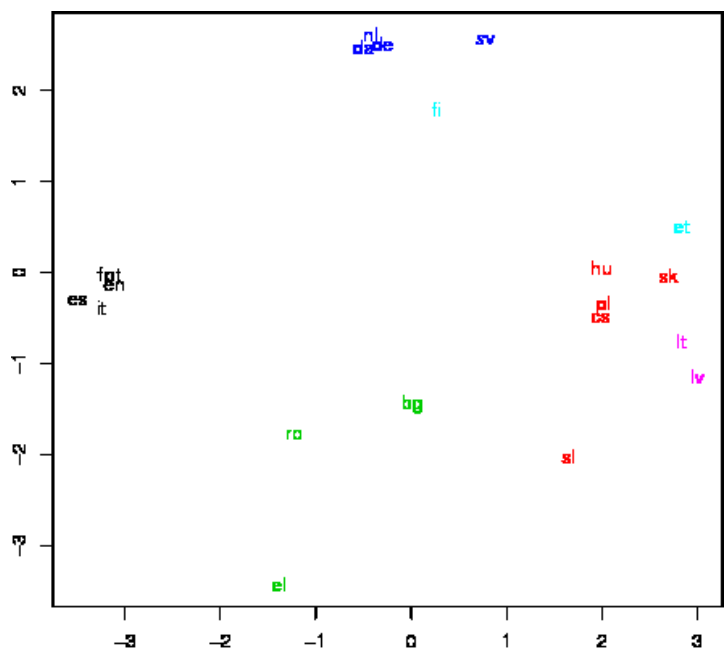}
  \caption{Two-dimensional PCA (principal component analysis)
  projection of average distance matrices 
  as described in the text.}
  \label{fig:all21}
\end{center}
\end{figure}
\begin{table}
\begin{center}
{ 
\begin{tabular}{|c|lll|} \hline
       &  Geo & WALS & Sem \\ \hline
Geo &  & 5\%/45\%** & 40\%***/65\%*** \\
WALS & & & 23\%**/62\%***\\
Sem & & & \\ \hline
\end{tabular}
}
\caption{{Correlation between dist.\ matrices, Mantel test, \texttt{All21/Big10}.}}
\label{table:distances}
\end{center}
\end{table}
Observing that fi is relatively similar to sv, which is at odds with
genealogical/structural language classifications, we test another 
question, namely, whether the resulting semantic 
distance matrix is more similar to a distance matrix based on
genealogical/structural relationships or to a distance matrix based on
geographic relations. To this end, we 
determine the degree of
structural similarity between two languages as the number of agreeing
features (a feature is, e.g., \textit{Number of Cases}) in the WALS\footnote{{http://wals.info/}} database of structural
properties of languages divided by the number of total features available for
the language pair \cite{Cysouw_other:2013}. For geographic distance, we use
the dataset from \newcite{CEPII:2011-25} which lists distances between countries. We
make the simplifying assumption that, e.g., language it and country
Italy agree, i.e., it is spoken in Italy (exclusively). Table
\ref{table:distances} shows that geographic distance correlates better
with our semantic distance calculation than does WALS structural
similarity under the Mantel test measure.
This may hint at
an interesting result: since   
semantics is 
changing fast, 
it may be more directly influenced by contact phenomena than by
genealogical processes 
that operate 
on a much slower
time-scale. 

{Note that our results are in accordance with the assumption that the probability of borrowing and geographical distance are inversely correlated \cite{Cysouw:2013}.
In our case, this may relate to semantic loans (adopting the semantic neighborhoods of loaned words within the target language) rather than to structural or grammatical borrowings.
That is, geographically related languages exhibit a higher probability to borrow words 
from each other 
together with the same range of semantic associations.
At least, this hypothesis is not falsified by our experiment.}

\section{Discussion}\label{sec:discussion}
Our initial expectation was that `distant' second languages $\ell$ --- in
terms of language similarity --- would greatly deteriorate monolingual
semantic evaluations in a target language $p$, as we believed they
would invoke 
`unusual' semantic associations from the perspective of $p$. Such a
finding would have been a word of caution regarding with which
language to embed a target language $p$ in a joint semantic space, if
this happens for the sake of improving monolingual semantic
similarity in $p$. 
We
were surprised to find that only BBA was 
sensitive to language
similarity in our experiments in this respect, 
whereas CCA seems quite robust
against choice of second language. An explanation for this finding may
be the different manners in which both methods induce joint embedding
spaces: While CCA takes independently constructed vector spaces of
two languages, BBA jointly optimizes mono- and bilingual constraints
and may thus be more sensitive to the interplay, and relation, between
both languages. 
Another plausible explanation is that CCA uses
only \emph{1-best} alignments for projecting two languages in a joint
semantic space. Thus, it 
may be 
less sensitive to varying polysemous associations across
different languages (cf.\ our \emph{vir} example in Section \ref{sec:introduction}), 
and hence 
less adequate for 
capturing cross-lingual polysemy.\footnote{Thus, we would also expect
CCA to perform better in monolingual intrinsic evaluations (as our
experiments have partly confirmed) and BBA to perform better in
multilingual intrinsic evaluations. We thank one reviewer for
pointing this out.} 

In terms of language similarity, we mention that our approach is
formally 
similar to approaches as
in \cite{Eger:2015,Asgari:2016} and others. Namely, we construct
graphs, one for each language, and 
compare 
them 
to determine language distance. 
Compared to \newcite{Eger:2015}, our approach differs in that they use
translations in a second language $\ell$ to measure similarity between
pivot language $p$ words. 
This 
idea also 
underlies very well-known
lexical semantic resources such as the paraphrase database (PPDB)
\cite{Bannard:2005,Ganitkevitch:2013}; see also \newcite{Eger:2010}. In
contrast, we directly use 
bilingual embeddings for this similarity measurement by jointly
embedding $p$ and $\ell$, which are arguably best suited for this
task. Our approach also differs from 
\newcite{Eger:2015}
in that
we do not apply a random-surfer process to our semantic graphs. 

We finally note that the
linguistic problem of (semantic) language classification, as we
consider, involves some vagueness as there is de facto no gold
standard that we can compare to. 
Reasonably, however, 
languages should be semantically similar to a degree that reflects
structural, genealogical, and contact relationships. One approach may
then be to disentangle or, as we pursued here, (relatively) weigh 
each of these effects.    

{From an application perspective, our approach allows for enriching (automatically generated) lexica. 
This relates, for example, to the generation of sentiment lexica listing prior polarities for selected lexemes \cite{Sonntag:Stede:2015}. 
%
%
Since the acquisition of such specialized information (e.g., by annotation) is cost-intensive, approaches are needed that allow for automatically generating or extending such resources especially in the case of historical languages (e.g., Latin).
Here our idea is to start from pairs of semantically (most) similar languages in order to induce polarity cues for words in the target language as a function of their distances to selected seed words in the pivot language, for which polarities are already known.
By averaging over groups of semantically related pivot languages, for which sentiment lexica already exist, the priority listings for the desired target language may stabilize.
Obviously, this procedure can be applied to whatever lexical information to be annotated automatically (e.g., grammatical or semantic categories like agency, animacy etc. as needed, for example, for semantic role labeling \cite{Palmer:Gildea:Xue:2010}).}

{A second application scenario relates to measuring (dis-)similarities of translations and their source texts \cite{Baker:1993}: 
starting from our model of bilingual semantic spaces, we may ask, for example, whether words for which several alternatives exist within the target language tend to be translated by candidates that retain most of their associations within the source language -- possibly in contradiction to frequency effects.
%
%
Such a finding would be in line with Toury's notion of interference \cite{Toury:1995} according to which translations reflect characteristics of the source language -- the latter leaves, so to speak, fingerprints within the former.
Such a finding would bridge between the notion of interference in translation studies and distributional semantics based on deep learning.}

\section{Related work}\label{sec:related}
Besides the mono- and multilingual word vector representation research 
that forms the basis of our work and which has already been referred
to, we mention the following three related approaches to language
classification. \newcite{Koehn:2005} compares down-stream task
performance in SMT to language family relationship, finding positive
correlation. \newcite{Cooper:2008} measures semantic language distance
via bilingual dictionaries, finding that French appears to be
semantically closer to Basque than to German, supporting our arguments
on contact as co-determining semantic language similarity. 
\newcite{Bamman:2014} and \newcite{Kulkarni:2015} study semantic
distance between dialects of  
 English by comparing region specific word embeddings. 

Studying geographic variation of (different) languages is also closely
related to 
studying temporal variation within one and the same
language \cite{Kulkarni:2015b}, with one crucial difference being the need
to find a common representation in the former
case. Word embeddings --- in particular, monolingual ones --- can also
be used to 
address the latter scenario \cite{Eger:2016,Hamilton:2016}.     

In terms of classifying languages, the work that is closest to ours is
that of \newcite{Asgari:2016}. A 
key difference between their approach and ours is that, in order to
achieve a common representation between languages, they translate
words. This has the disadvantage that translation pairs need to be
known, which typically requires large amounts of parallel text. In
contrast, bilingual word embeddings, which form the basis of our
experiments,  
can be generated from as few as ten
translation 
pairs, as demonstrated in \newcite{Zhang:2016}. 

There is by now a long-standing tradition that compares languages
via analysis of complex networks that encode their words 
and the (semantic) relationships between
them \cite{Cancho:2001,Gao:2014}. These studies often only
look at very abstract statistics of networks such as average path
lengths and clustering coefficients, rather than analyzing them on
a level of content of their nodes and edges.  
In addition, they often
substitute co-occurrence as a proxy for semantic similarity. However,
as \newcite{Asgari:2016} point out, co-occurrence is a naive estimate
of similarity; e.g., synonyms rarely co-occur. 


\section{Conclusion}\label{sec:conclusion}
%
Using English, German and Italian as pivot languages, we show that the
choice 
of the second language may significantly matter when the resulting
space is used for  
monolingual semantic evaluation tasks. 
More specifically, we show that the goodness of this choice is
influenced by genealogical similarity and by (geographical) language
contact.
This finding may be important for the question which languages to
integrate in multilingual embedding spaces \cite{Huang:2015}. 
%
Moreover, we show 
that semantic language similarity --- estimated on the basis of bilingual
embedding spaces as suggested in this work --- may be better predicted by
contact than by genealogical relatedness.  
%
%
The validation of this hypothesis by means of bigger data sets 
will be the object of future
work. 
%


\bibliography{acl2016}
\bibliographystyle{acl2016}

\appendix

\end{document}